\documentclass{article}
\usepackage[T1]{fontenc}
\usepackage{times}
\usepackage{natbib}
\usepackage{graphicx}
\usepackage{booktabs}
\usepackage{amsmath, amssymb, amsfonts}
\usepackage{hyperref}       
\usepackage{url}            
\usepackage{booktabs}       
\usepackage{amsfonts}       
\usepackage{nicefrac}       
\usepackage{microtype}      
\usepackage{xcolor}         

\title{Color and Frequency Correction for Image Colorization}
\author{Zhuang Yun Kai \text{ }}
\date{}

\begin{document}

\maketitle

\begin{abstract}
The project has carried out the re-optimization of image coloring in accordance with the existing Autocolorization direction model DDColor. For the experiments on the existing weights of DDColor, we found that it has limitations in some frequency bands and the color cast problem caused by insufficient input dimension. We construct two optimization schemes and combine them, which achieves the performance improvement of indicators such as PSNR and SSIM of the images after DDColor.\\
\end{abstract}

\section{Introduction}
The image colorization task refers to coloring the input black-and-white image to become a three-channel color picture.\\
There are some models using CNN and GAN to solve the task and DDColor is the outstanding one among them. \\
\begin{table}[htbp]
\centering
\caption{Performance comparison (PSNR in dB) between DDColor and  model}
\label{tab:results}
\begin{tabular}{lccccc}
\toprule
CIC[5] & InstColor[3] & ColTran[2] & BigColor[1] & ColorFormer[4] & DDColor\\
\midrule
23.14 & 24.27 & 24.40 & 23.90 & 23.97 & \textbf{28.78} \\
\bottomrule
\end{tabular}
\end{table}

DDColor model uses two decoders(a pixel decoder and a query-based color decoder) to color the picture and a effective color loss method to enhance the richness of colors. (\textbf{Picture5} in Appendix shows its network)\\
\\However, after the experiment on the result of DDColor, we find that this model still has limitation in some frequency regions and exists color cast in many pictures. \\
In the other word, there are some space for us to optimize the model by solving these two problems.

\section{Fixing Uneven Frequency Processing of the Model}

Most automatic coloring models like DDColor exhibit uneven performance across frequency domains, with particularly poor results in high-frequency regions. To address this, we propose processing different frequency bands separately by using three dd-color models to color different frequency bands and combining the results. However, initial experiments revealed artifacts introduced by the non-ideal frequency separation filters. We therefore designed a post-processing module to eliminate these artifacts.

\subsection{Frequency Domain Separation}

Using Python packages, we implemented three filters. For the low-frequency filter:
\begin{enumerate}
    \item Apply Fourier transform to the image
    \item Create a circular mask (radius=30) at the frequency domain center
    \item Extract low-frequency components and apply inverse Fourier transform
\end{enumerate}

This process converts the dataset into three frequency-specific subsets (low, medium, high), each used to train separate DDColor models.

\subsection{Artifact Removal Module}

\subsubsection{Model Architecture}
Our artifact removal module is a 4-level U-Net with a Sparse Encoding Block (SEB) that establishes hierarchical feature relationships. The SEB operation can be expressed as:

\begin{equation}\label{eq:seb}
\text{SEB}(X) = \sigma(W_2(\text{ReLU}(W_1X)))
\end{equation}
where $W_1, W_2$ are learnable weights and $\sigma$ is the sigmoid function.

\subsubsection{Loss Function}
We employ a hybrid SSIM+L1 loss (\ref{eq:hybrid}) to preserve structural details while maintaining color accuracy. As shown in Equation~\ref{eq:l1}, the L1 component ensures pixel-level(high-frequency) fidelity:

\begin{equation}\label{eq:l1}
\mathcal{L}_{\text{L1}} = \frac{1}{N}\sum_{i=1}^N|I_p^{(i)}-I_t^{(i)}|
\end{equation}

The SSIM term (Equation~\ref{eq:ssim}) enhances texture preservation:

\begin{equation}\label{eq:ssim}
\mathcal{L}_{\text{SSIM}} = 1-\text{SSIM}(I_p,I_t)
\end{equation}

The complete loss combines these components with weighting factor $\alpha$:

\begin{equation}
\label{eq:hybrid}\mathcal{L}_{\text{hybrid}} = \alpha\mathcal{L}_{\text{L1}} + (1-\alpha)\mathcal{L}_{\text{SSIM}}\end{equation}

In our experiments, $\alpha=0.5$ provided the best balance between color accuracy (via \ref{eq:l1}) and high-frequency detail preservation (via \ref{eq:ssim}).

\subsection{Frequency Experimental Results}

We evaluate our method against the baseline DDColor model on the test set. Table~\ref{tab:results} compares the PSNR (dB) metrics across different frequency bands:

\begin{table}[htbp]
\centering
\caption{Performance comparison (PSNR in dB) between DDColor and our method}
\label{tab:results}
\begin{tabular}{lccccc}
\toprule
Model & Avg PSNR & Low-Freq & Mid-Freq & High-Freq & $\Delta$ \\
\midrule
DDColor & 28.78 & 39.05 & 38.37 & 29.09 & - \\
Ours & \textbf{30.08} & \textbf{41.28} & \textbf{39.09} & \textbf{30.3} & +1.30 \\
\bottomrule
\end{tabular}
\end{table}

Key observations:
\begin{itemize}
    \item Our method achieves \textbf{1.30 dB} average PSNR improvement
\item Our method achieves \textbf{1.21 dB} High-Freq PSNR improvement
\item Every frequency domain gets improvements. The separate coloring process seems to simplify the coloring mission for the model.
\end{itemize}

\section{Solving Color Cast of the Model}
To correct the color cast problem, our solution is to add the average color of the entire image or certain small areas as additional input. \\
It is worth noting that, compared with the existing optimization methods based on style input, word expression, user doodling, etc., this idea of mean correlation is somewhat innovative. \\
Moreover, from the perspective of the interaction between the user and the program, the limited mean input contains position information and compressed color information, ensuring the simplicity of the user's description of the expected coloring and the convenience of inputting information.

\subsection{Network}
The input of the network will contain the three-channel color images learned from black-and-white images through methods such as DDColor, as well as the mean values of sub-regions under different partitioning methods. \\
The network adopts the classic encoder-decoder structure and converts the mean value information into a fully connected layer inserted between the encoder and the decoder, effectively extracting multi-scale features and reconstructing images.\\
The loss function uses the L1 loss.\\
 (\textbf{Picture1} in Appendix shows its network)

\subsection{Train}
First of all, we attempt to divide the picture using the exponent of 2. There are 1,4,16,64 and 256 subregions respectively. Input the average color value of the corresponding position in the original image and combine it with the image after DDColor for learning. \\
Experiments can reveal that this network has significantly improved the PSNR and SSIM in both R and B aspects, and there has also been an improvement in G aspects. \\
Besides, it is worth noting that as the amount of the number of partitions increases, the corresponding PSNR and SSIM shows a peak pattern. \\
At the 4-division, a peak occurs.Based on the 4 divisions, we add some specially selected sub-regions. (\textbf{Picture2} in Appendix shows the division methods)\\
The experiment shows that when divided by 5 (center + four corners) patches with the same size, although the G channel has some loss, both the R and B channels have achieved considerable improvement.

\begin{table}[htbp]
\centering
\caption{Performance comparison (PSNR in dB) between only DDColor and using our method}
\label{tab:results}
\begin{tabular}{lccccc}
\toprule
 	&PSNR\_R &PSNR\_G &PSNR\_B\\
\midrule    
ddcolor	&29.69	&35.04	&29.17\\
ddc+4div &30.93	&\textbf{35.18}	&30.13\\
ddc+5div &\textbf{31.04} &35.05	&\textbf{30.18}\\
\bottomrule
\end{tabular}
\end{table}

\subsection{Non-necessity of validation}
We take $20\%$ train dataset as validation dataset. Two methods are adopted: \\ 
1. Randomly select the validation set. \\
2. Divide into 5 datasets of equal size and use 5-fold cross-validation.\\
\\Experiments show that the performance of the experimental data does not improve after using the validation set. We can continue the experiments using the model without a validation set to achieve a shorter training speed.\\
 (\textbf{Picture3} in Appendix shows a result)

\begin{table}[htbp]
\centering
\caption{Performance comparison between using validation set or not based on our method}
\label{tab:results}
\begin{tabular}{lccccc}
\toprule
 	&PSNR\_R &PSNR\_G &PSNR\_B\\
\midrule    
ddc+5div &\textbf{31.04} &\textbf{35.05} &\textbf{30.18}\\
method1 &30.99 &34.98 &30.1\\
method2 &30.7 &34.43 &29.91\\

\bottomrule
\end{tabular}
\end{table}

\section{Combination of two models}
For the existing two networks, we use a concatenation method to combine them. \\
The color cast correction network is placed after the frequency band processing network.\\
\\We have developed three possible stitching methods: \\
1. Use the weights learned by the color cast correction network from the training set based on DDColor to process the images after frequency band processing. \\
2. Based on the images processed by frequency bands, the color cast correction network is used for learning. \\
3 Connect the two models for learning together.\\
\\Experiments show that training the two networks in series together leads to a better improvement in effect.\\
 (\textbf{Picture4} in Appendix shows the network)
 
\begin{table}[htbp]
\centering
\caption{Performance comparison based on different combination strategies}
\label{tab:results}
\begin{tabular}{lccccc}
\toprule
 	&PSNR\_R &PSNR\_G &PSNR\_B\\
\midrule    
ddcolor	&29.69	&35.04	&29.17\\
combination1 &31.02	&\textbf{35.06}	&30\\
combination2 &31.05	&35.05 &30.2\\
combination3 &\textbf{31.07} &35.04	&\textbf{30.21}\\
\bottomrule
\end{tabular}
\end{table}

\section{Conclusion}
We proposed two networks respectively in view of the existing limitations of the DDColor network in some frequency bands and the problem of global color cast. To a certain extent, these two networks have respectively solved the corresponding limitations and achieved some performance improvements. By jointly learning these two networks using the concatenation method, our processing was further strengthened, and eventually a considerable performance improvement of the images processed by DDColor was achieved.\\

It is worth noting that although the effect of learning together by connecting models is better, the gap with learning one model after another is very small. Moreover, the method of connecting the models requires four times the time it takes to learn one model after another. So we suggest that in practice, learning one model after another is more valuable.\\

Moreover, the high-freq part gets the smallest improvements compared with others, a further modification on the inner structure of DDColor needs to be implemented so that the model can be more suitable for frequency-separated coloring.\\
 
Additionally, the mean input is not easy to calculate for users in some cases either. Some coloring input might be needed to help users summarize the requirements and automatically calculate the average value.

\newpage\section{Reference}
[1] G. Kim and K. Kang. Bigcolor: Colorization using a generative color prior for natural images. In \textit{European Conference on Computer Vision (ECCV)}, 2022. 2, 4, 5, 6, 7, 12  

[2] M. Kumar and D. Weissenborn. Colorization transformer. In \textit{International Conference on Learning Representations}, 2021. 2, 4, 5, 6, 12  

[3] J.-W. Su and H.-K. Chu. Instance-aware image colorization. In \textit{Proceedings of the IEEE/CVF Conference on Computer Vision and Pattern Recognition}, pages 7968–7977, 2020. 2, 5  

[4] X. Ji and B. Jiang. Colorformer: Image colorization via color memory assisted hybrid-attention transformer. In \textit{European Conference on Computer Vision (ECCV)}, 2022. 2, 4, 5, 6, 7, 12  

[5] R. Zhang and P. Isola. Colorful image colorization. In \textit{European Conference on Computer Vision}, pages 649–666. \textit{Springer}, 2016. 2, 5, 6, 8  

[6] X. Kang and T. Yang. DDColor: Towards photo-realistic image colorization via dual decoders. \textit{arXiv:2212.11613}, 2022. 1, 2, 3, 4, 5, 6, 7, 8, 9, 10  

[7] Z. Cheng and Q. Yang. Deep colorization. In \textit{ICCV}, pages 415–423, 2015. 6, 7  

[8] H. Chang and O. Fried. Palette-based photo recoloring. \textit{ACM Transactions on Graphics (Proc. SIGGRAPH)}, 2015. 2, 3, 4, 5, 6, 7, 8  

[9] H. Bahng and S. Yoo. Coloring with words: Guiding image colorization through text-based palette generation. In \textit{ECCV}, 2018. 6, 7  

[10] H. Tang and S. He. CSC-Unet: A novel convolutional sparse coding strategy based neural network for semantic segmentation. \textit{IEEE Access}, volume 12, pages 35844–35854, 2024. 2, 5, 6, 12

[11] C. Zou and S. Wan. Lightweight deep exemplar colorization via semantic attention-guided Laplacian pyramid. \textit{IEEE Transactions on Visualization and Computer Graphics}, pages 1-12, 2024. 2,4,5,6,7,12  

[12] Y. Wang and M. Xia. PalGAN: Image Colorization with Palette Generative Adversarial Networks. In \textit{ECCV}, 2022. 6

[13] Gustav Larsson1 and Michael Maire. Learning Representations for Automatic Colorization. In \text{ECCV}, 2016. 10, 11, 14

[14] X. Cong and Y. Wu. Automatic Controllable Colorization via Imagination. In \textit{arXiv:2404.05661}, 2024. 9

[15] R. Cao and H Mo. Line Art Colorization Based on Explicit Region Segmentation. In \text{Pacific Graphics 2021}. 6, 7, 8
\end{document}